\documentclass[letterpaper,10pt,conference]{ieeeconf}

\IEEEoverridecommandlockouts

\usepackage[export]{adjustbox}
\usepackage{amssymb}
\usepackage{bm}
\usepackage{booktabs,arydshln}
\usepackage{glossaries}
\usepackage[bookmarks=true]{hyperref}
\usepackage{multirow}
\usepackage{rotating}
\usepackage[binary-units=true]{siunitx}
\usepackage{subcaption}

\newacronym{DoF}{DoF}{degrees of freedom}
\newacronym{MRP}{MRP}{modified Rodrigues parameters}
\newacronym{NLP}{NLP}{nonlinear programming}
\newacronym{RBD.jl}{RBD.jl}{RigidBodyDynamics.jl}
\newacronym{RMSE}{RMSE}{root-mean-square error}
\newacronym{wrt}{w.r.t.}{with respect to}

\graphicspath{{figures/}}

\hypersetup{hidelinks}

\makeatletter
\def\adl@drawiv#1#2#3{%
    \hskip.5\tabcolsep
    \xleaders#3{#2.5\@tempdimb#1{0.5}#2.5\@tempdimb}%
             #2\z@ plus1fil minus1fil\relax
    \hskip.5\tabcolsep}
\newcommand{\cdashlinelr}[1]{%
    \noalign{\vskip\aboverulesep
             \global\let\@dashdrawstore\adl@draw
             \global\let\adl@draw\adl@drawiv}
    \cdashline{#1}
    \noalign{\global\let\adl@draw\@dashdrawstore
             \vskip\belowrulesep}}
\makeatother

\hypersetup{
    pdfauthor   = {Henrique Ferrolho},%
    pdftitle    = {Inverse Dynamics vs. Forward Dynamics in Direct Transcription Formulations for Trajectory Optimization},%
    pdfsubject  = {Robotics},%
    pdfkeywords = {Robots, Trajectory Optimization, Direct Transcription, Dynamics}%
}

\title{\LARGE\bf
Inverse Dynamics vs. Forward Dynamics \\ in Direct Transcription Formulations for Trajectory Optimization
}

\author{
    Henrique Ferrolho${}^{1}$, Vladimir Ivan${}^{1}$, Wolfgang Merkt${}^{2}$, Ioannis Havoutis${}^{2}$, Sethu Vijayakumar${}^{1}$%
    \thanks{${}^{1}$School of Informatics, University of Edinburgh, United Kingdom.}%
    \thanks{${}^{2}$Oxford Robotics Institute, University of Oxford, United Kingdom.}%
    \thanks{
        This research is supported by
        EPSRC UK RAI Hub for Offshore Robotics for Certification of Assets (ORCA, EP/R026173/1),
        EU H2020 project Memory of Motion (MEMMO, 780684), and
        EPSRC as part of the Centre for Doctoral Training in Robotics and Autonomous Systems at Heriot-Watt University and The University of Edinburgh (EP/L016834/1).
    }%
    \thanks{\textit{Email address:} {\href{mailto:henrique.ferrolho@ed.ac.uk}{\nolinkurl{henrique.ferrolho@ed.ac.uk}}}}%
}

\begin{document}
\bstctlcite{IEEEexample:BSTcontrol}

\maketitle
\thispagestyle{empty}
\pagestyle{empty}

\begin{abstract}
    Benchmarks of state-of-the-art rigid-body dynamics libraries report better performance solving the inverse dynamics problem than the forward alternative.
    Those benchmarks encouraged us to question whether that computational advantage would translate to direct transcription, where calculating rigid-body dynamics and their derivatives accounts for a significant share of computation time.
    In this work, we implement an optimization framework where both approaches for enforcing the system dynamics are available.
    We evaluate the performance of each approach for systems of varying complexity, for domains with rigid contacts.
    Our tests reveal that formulations using inverse dynamics converge faster, require less iterations, and are more robust to coarse problem discretization.
    These results indicate that inverse dynamics should be preferred to enforce the nonlinear system dynamics in simultaneous methods, such as direct transcription.
\end{abstract}

\section{Introduction}
\label{sec:introduction}

Direct transcription \cite{betts2010practical} is an effective approach to formulate and solve trajectory optimization problems.
It works by converting the original trajectory optimization problem (which is \emph{continuous} in time) into a numerical optimization problem that is \emph{discrete} in time, and which in turn can be solved using an off-the-shelf \gls{NLP} solver.
First, the trajectory is divided into segments and then, at the beginning of each segment, the system state and control inputs are explicitly discretized---these are the decision variables of the optimization problem.
Due to this discretization approach, direct transcription falls under the class of \emph{simultaneous} methods.
Finally, a set of mathematical constraints is defined to enforce boundary and path constraints, e.g., initial and final conditions, or intermediate goals.
In dynamic trajectory optimization, there exists a specific set of constraints dedicated to enforce the equations of motion of the system, the so-called \emph{defect constraints}.
This paper discusses different ways of defining these constraints, as well as their implications.

The dynamics defects are one of the most important constraints in optimization problems when planning highly dynamic motions for complex systems, such as legged robots.
Satisfaction of these constraints ensures that the computed motion is reliable and physically consistent with the nonlinear dynamics of the system.
The dynamics defect constraints are usually at the very core of optimal control formulations, and require computing rigid-body dynamics and their derivatives---which account for a significant portion of the optimization computation time.
Therefore, it is of utmost importance to use an algorithm that allows to compute the dynamics of the system reliably, while achieving low computational time.

\begin{figure}[t]
    \captionsetup{font=small}
    \begin{subfigure}[t]{0.5\linewidth}\includegraphics[width=0.99\linewidth,left]{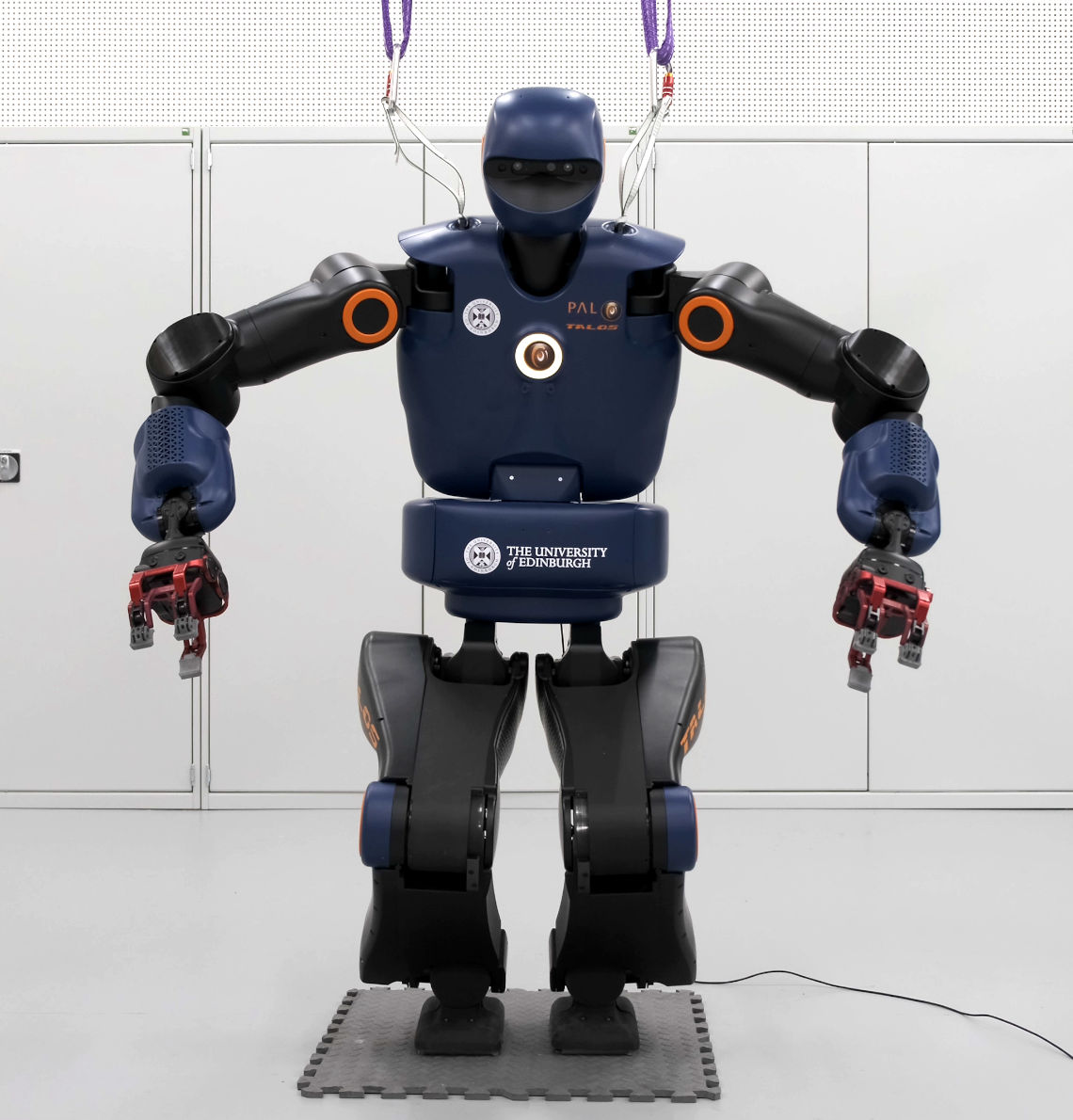}\end{subfigure}\hfill%
    \begin{subfigure}[t]{0.5\linewidth}\includegraphics[width=0.99\linewidth,right]{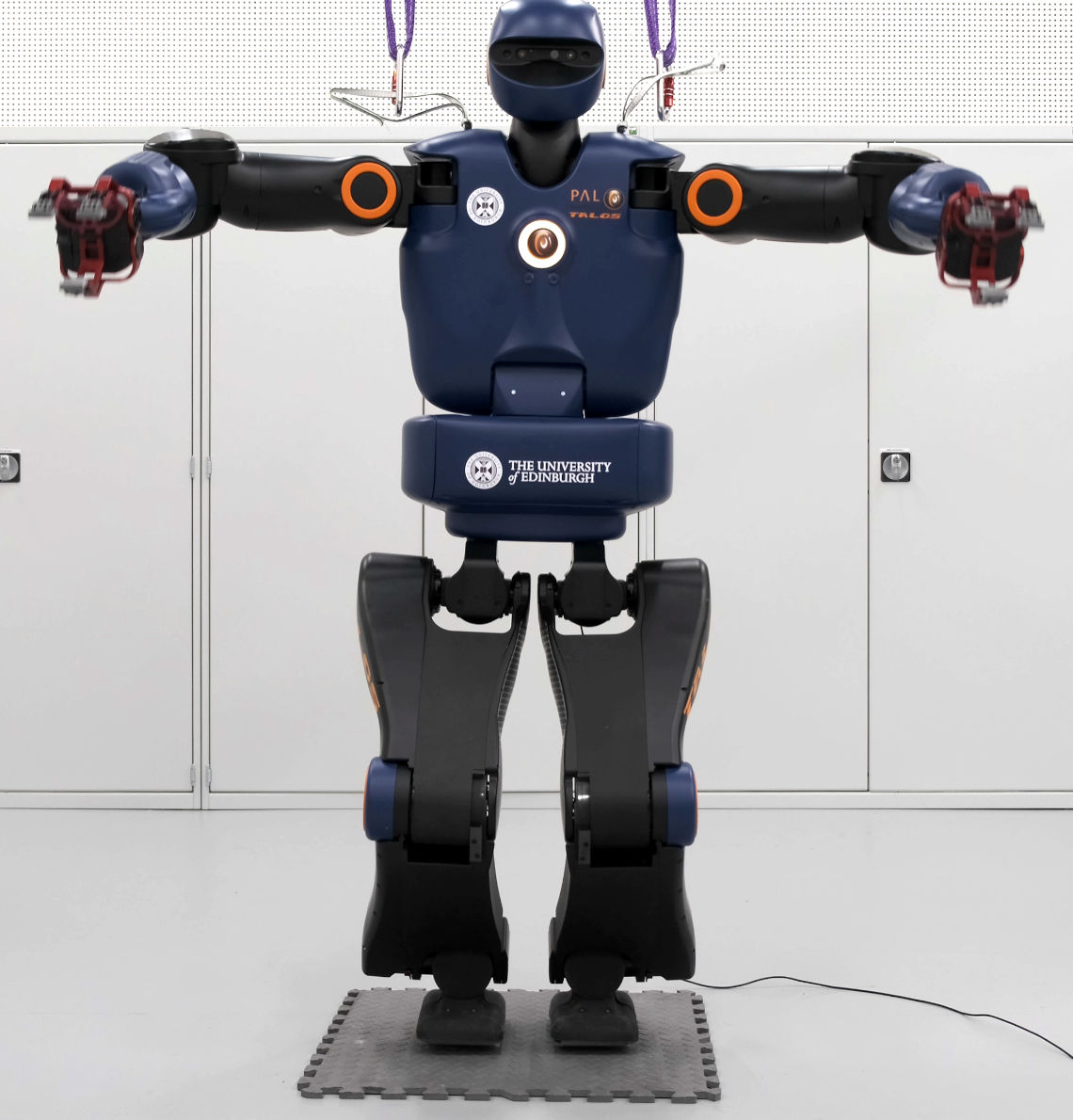}\end{subfigure}\hfill%
    \vspace{0.01\linewidth}
    \begin{subfigure}[t]{\linewidth}\includegraphics[width=\linewidth,center]{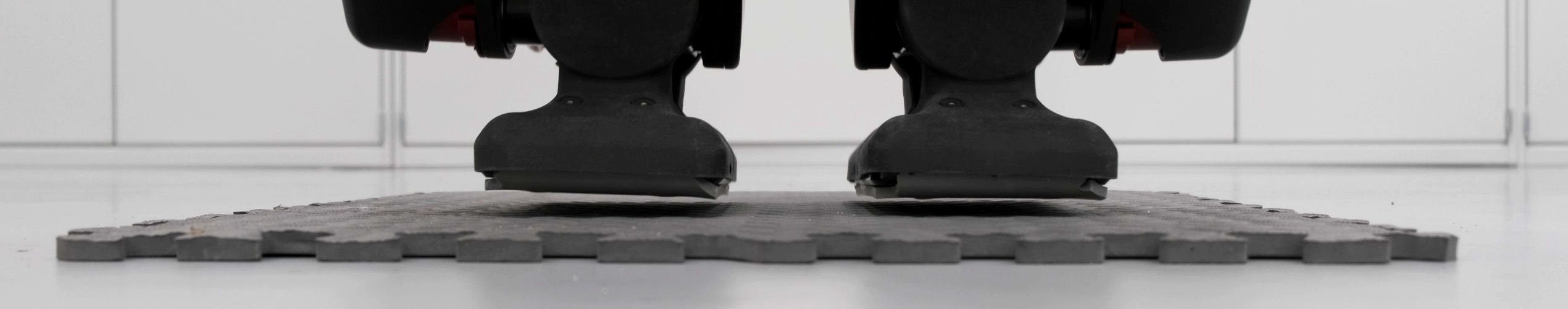}\end{subfigure}
    \caption{Snapshots of the humanoid TALOS~\cite{stasse2017talos} jumping.}\label{figure:cover}
    \vspace{-12pt}
\end{figure}

In the study of the dynamics of open-chain robots, the \emph{forward dynamics} problem determines the joint accelerations resultant from a given set of joint forces and torques applied at a given state.
On the other hand, the \emph{inverse dynamics} problem determines the joint torques and forces required to meet some desired joint accelerations at a given state.
In trajectory optimization, most direct formulations use forward dynamics to enforce dynamical consistency \cite{pardo2016evaluating}.
However, benchmarks have shown that most dynamics libraries solve the \emph{inverse dynamics} problem (e.g., with the Recursive Newton-Euler Algorithm) faster than the \emph{forward dynamics} problem (e.g., with the Articulated Body Algorithm) \cite{koolen2019julia,neuman2019benchmarking}.
For example, for the humanoid robot TALOS~\cite{stasse2017talos}, the library Pinocchio~\cite{carpentier2019pinocchio} solves the inverse dynamics problem in just \SI{4}{\micro\second}, while the forward dynamics problem takes \SI{10}{\micro\second}.
These differences in performance motivated us to question whether the computational advantage of inverse dynamics would translate to direct transcription---where the dynamics problem needs to be solved several times while computing the defect constraints.
Moreover, there is biological evidence suggesting that inverse dynamics is employed by the nervous system to generate feedforward commands~\cite{schweighofer1998role}, while other studies support the existence of a forward model~\cite{mehta2002forward}---which increased our interest in this topic.

In this work, we present a trajectory optimization framework for domains with rigid contacts, using a direct transcription approach.
Particularly, our formulation allows to define dynamics defect constraints employing either forward dynamics or inverse dynamics.
We defined a set of evaluation tasks across different classes of robot platforms, including fixed- and floating-base systems, with point and surface contacts.
Our results showed that inverse dynamics leads to significant improvements in computational performance when compared to forward dynamics---supporting our initial hypothesis.

\section{Related Work}
\label{sec:related_work}

\gls{RBD.jl}~\cite{rigidbodydynamicsjl}, RBDL~\cite{felis2016rbdl}, Pinocchio~\cite{carpentier2019pinocchio}, and RobCoGen~\cite{frigerio2016robcogen} are all state-of-the-art software implementations of key rigid-body dynamics algorithms.
Recently, Neuman \textit{et al.} \cite{neuman2019benchmarking} benchmarked these libraries and revealed interesting trends.
One such trend is that implementations of inverse dynamics algorithms have faster runtimes than forward dynamics.\footnote{%
    RobCoGen is an exception to this observation as it implements a hybrid dynamics solver which has a higher computational cost, and is significantly different from the implementations used by the other libraries.
}
Koolen and Deits \cite{koolen2019julia} also compared \gls{RBD.jl} with RBDL, and their results showed that solving inverse dynamics was at least two times faster than solving forward dynamics for the humanoid robot Atlas.
Both of these studies only consider computation time of rigid-body dynamics; they do not provide insight into how these algorithms perform when used in trajectory optimization.

Lee \textit{et al.} \cite{lee2005newton} have proposed Newton and quasi-Newton algorithms to optimize motions for serial-chain and closed-chain mechanisms using inverse dynamics.
However, they used relatively simple mechanisms for which analytic derivatives can be obtained.
In this work, we are interested in dynamic motions of complex mechanisms in domains with contact, for which the derivation of analytic derivatives is an error-prone process, involving significant effort.

In the same spirit, Erez and Todorov \cite{erez2012trajectory} generated a running gait for a humanoid based on inverse dynamics under external contacts.
This method allowed them to formulate an \emph{unconstrained} optimization where all contact states can be considered equally, contact timings and locations are optimized, and reaction forces are computed using a smooth and invertible contact model \cite{todorov2011convex} with convex optimization.
However, their approach requires ``helper forces'', as well as tuning of contact smoothness and of the penalty parameters on the helper forces to achieve reasonable-looking behavior.
In contrast, our approach does not require helper forces or any tuning whatsoever; we consider contact forces as decision variables and model contacts rigidly.
Another difference is that we formulate a \emph{constrained} optimization problem and enforce the nonlinear system dynamics with hard constraints, which results in high-fidelity motions.
This is especially important for deployment on real hardware, where dynamic consistency and realism are imperative.
The main focus of our paper is not the contact problem, and we assume contact locations and contact times are known \textit{a priori}.

Finally, to the best of our knowledge, there is no current work directly comparing inverse dynamics against forward dynamics in the context of direct methods.
Posa \textit{et al.} \cite{posa2014direct} also identified that a formal comparison is important, but missing so far.
They argued that one of the reasons for this was that the field had not yet agreed upon a set of canonical and hard problems.
In this paper, we tackle this issue, and compare the two approaches on robots of different complexity on a set of dynamic tasks.

The main contributions of this work are:
\begin{enumerate}
    \item A direct transcription formulation that uses \emph{inverse dynamics} to enforce physical consistency, for constrained trajectory optimization in domains with rigid contacts.
    \item Evaluation of the performance of direct transcription formulations using either forward or inverse dynamics, for different classes of robot platforms: a fixed-base manipulator, a quadruped, and a humanoid.
    \item Comparison of performance for different linear solvers, and across strategies to handle the barrier parameter of the interior point optimization algorithm.
\end{enumerate}
We validated our trajectories in full-physics simulation and with hardware experiments.
We also open-sourced a version of our framework for fixed-base robots, TORA.jl~\cite{torajl}.

\section{Trajectory Optimization}
\label{sec:trajectory_optimization}

\subsection{Robot Model Formulation}
We formulate the model of a legged robot as a free-floating base $B$ to which limbs are attached.
The motion of the system can be described \gls{wrt} a fixed inertial frame $I$.
We represent the position of the free-floating base \gls{wrt} the inertial frame, and expressed in the inertial frame, as ${}_I\bm{r}_{IB} \in \mathbb{R}^3$;
and the orientation of the base as $\bm{\psi}_{IB} \in \overline{\mathbb{R}}{}^3$, using \gls{MRP}~\cite{gormley1945stereographic,terzakis2018modified}.
The joint angles describing the configuration of the limbs of the robot (legs or arms) are stacked in a vector $\bm{q}_j \in \mathbb{R}^{n_j}$, where $n_j$ is the number of actuated joints.
The generalized coordinates vector $\bm{q}$ and the generalized velocities vector $\bm{v}$ of this floating-base system may therefore be written as
\begin{equation}
    \bm{q} = \begin{bmatrix} {}_I\bm{r}_{IB} \\ \bm{\psi}_{IB} \\ \bm{q}_j \end{bmatrix} \in \mathbb{R}^3 \times \overline{\mathbb{R}}{}^3 \times \mathbb{R}^{n_j}, \quad
    \bm{v} = \begin{bmatrix} \bm{\nu}_B \\ \bm{\dot{q}}_j \end{bmatrix} \in \mathbb{R}^{n_v},
\end{equation}
where the twist $\bm{\nu}_B = \left [ {}_I\bm{v}_{B} \quad {}_B\bm{\omega}_{IB} \right ]^\top \in \mathbb{R}^6$ encodes the linear and angular velocities of the base $B$ w.r.t. the inertial frame expressed in the $I$ and $B$ frames, and $n_v = 6 + n_j$.

For fixed-base manipulators, the generalized vectors of coordinates and velocities can be simplified to
$\bm{q} = \bm{q}_j \in \mathbb{R}^{n_j}$ and $\bm{v} = \bm{\dot{q}}_j \in \mathbb{R}^{n_j}$, due to the absence of a free-floating base.

\subsection{Problem Formulation}
We tackle the motion planning problem using trajectory optimization;
more specifically, using a \textit{direct transcription} approach.
The original problem is \emph{continuous} in time, so we start by converting it into a numerical optimization problem that is \emph{discrete} in time.
For that, we divide the trajectory into $N$ equally spaced segments,
$t_I = t_1 < \dots < t_M = t_F$,
where $t_I$ and $t_F$ are the start and final instants, respectively.
This division results in $M = N + 1$ discrete \textit{mesh points}, for each of which we explicitly discretize the states of the system, as well as the control inputs.
Let $x_k \equiv x(t_k)$ and $u_k \equiv u(t_k)$ be the values of the state and control variables at the $k$-th mesh point.
We treat $x_k \triangleq \{ \bm{q}_k, \bm{v}_k \}$ and $u_k \triangleq \{ \bm{\tau}_k, \bm{\lambda}_k \}$ as a set of \gls{NLP} variables, and formulate the trajectory optimization problem as
\begin{equation}
    \mathrm{find} \enspace \bm{\xi} \quad \mathrm{s.t.} \enspace x_{k+1} = f(x_k, u_k), \enspace x_k \in \mathcal{X}, \enspace u_k \in \mathcal{U},
    \label{equation:nlp}
\end{equation}
where $\bm{\xi}$ is the vector of decision variables,
$x_{k+1} = f(x_k, u_k)$ is the state transition function incorporating the nonlinear system dynamics,
and $\mathcal{X}$ and $\mathcal{U}$ are sets of feasible states and control inputs enforced by a set of equality and inequality constraints.
The decision variables vector $\bm{\xi}$ results from aggregating the generalized coordinates, generalized velocities, joint torques, and contact forces at every\footnote{The control inputs at the final state need not be discretized.} mesh point, i.e.,
\begin{align}
    \bm{\xi} \triangleq \{ \bm{q}_1, \bm{v}_1, \bm{\tau}_1, \bm{\lambda}_1, \cdots, \bm{q}_{N}, \bm{v}_{N}, \bm{\tau}_{N}, \bm{\lambda}_{N}, \bm{q}_M, \bm{v}_M \}.
\end{align}

Similarly to Winkler \textit{et al.} \cite{winkler2018gait} and differently to Erez and Todorov \cite{erez2012trajectory}, we transcribe the problem by only making use of hard constraints; and satisfaction of those constraints is a necessary requirement for the computed motions to be physically feasible and to complete the task successfully.
This design decision is motivated by the fact that considering a cost function requires expert knowledge to carefully tune the weighting parameters that control the trade-off between different objective terms.
Optimizing an objective function also requires additional iterations and computation time.
Nonetheless, for the sake of completion, one of the experiments we present later in this paper does include and discuss the minimization of a cost function.

For tasks where the robot makes or breaks contacts with the environment, we assume contact locations and contact timings are known \textit{a priori}.
This assumption allows us to enforce zero contact forces for mesh points where the robot is not in contact with the environment, and therefore our formulation does not require any actual complementarity constraints.
On the other hand, such assumption depends on pre-determined contact sequences specified either by a human or by a contact planner (such as \cite{winkler2018gait,tonneau2019sl1m,stouraitis2020online}).

\subsection{Problem Constraints}

\subsubsection{Bounds on the decision variables}
We constrain the joint positions, velocities, and torques to be within their corresponding lower and upper bounds.

\subsubsection{Initial and final joint velocities}
We enforce the initial and final velocities of every joint to be zero: $\bm{v}_1 = \bm{v}_M = \bm{0}$.

\subsubsection{End-effector pose}
We enforce end-effector poses with
$f^\mathrm{fk}(\bm{q}_k, i) = \bm{p}_i$,
where $f^\mathrm{fk}(\cdot)$ is the forward kinematics function,
$i$ refers to the $i$-th end-effector of the robot, and
$\bm{p}_i \in SE(3)$ is the desired pose.

\subsubsection{Contact forces}
For mesh points where the robot is \emph{not} in contact with the environment,
we enforce the contact forces at the respective contact points to be zero:
$\bm{\lambda}_k = \bm{0}$.

\subsubsection{Friction constraints}
We model friction at the contacts with linearized friction cones, in the same way as \cite{caron2015leveraging}.

\subsubsection{System dynamics}
We enforce nonlinear whole-body dynamics, $\dot{x} = f(x, u)$, with \emph{defect} constraints.
The approach used to define these constraints is the main subject of this paper,
and the next section explains this in detail.

\section{System Dynamics}
\label{sec:system_dynamics}
The equations of motion for a floating-base robot that interacts with its environment can be written as
\begin{equation}
    \bm{M}(\bm{q})\bm{\dot{v}} + \bm{h}(\bm{q}, \bm{v}) = \bm{S}^\top \bm{\tau} + \bm{J}_s^\top(\bm{q}) \bm{\lambda},
    \label{equation:equations_of_motion_floating_base}
\end{equation}
where $\bm{M}(\bm{q}) \in \mathbb{R}^{n_v \times n_v}$ is the mass matrix, and $\bm{h}(\bm{q}, \bm{v}) \in \mathbb{R}^{n_v}$ is the vector of Coriolis, centrifugal, and gravity terms.
On the right-hand side of the equation, $\bm{\tau} \in \mathbb{R}^{n_\tau}$ is the vector of joint torques commanded to the system,
and the selection matrix $\bm{S} = [\bm{0}_{n_\tau \times (n_v - n_\tau)} \quad \mathbb{I}_{n_\tau \times n_\tau}]$ selects which \gls{DoF} are actuated.
We consider that all limb joints are actuated, thus $n_\tau = n_j$.
The vector $\bm{\lambda} \in \mathbb{R}^{n_s}$ denotes the forces and torques experienced at the contact points, with $n_s$ being the total dimensionality of all contact wrenches.
The support Jacobian $\bm{J}_s \in \mathbb{R}^{n_s \times n_v}$ maps the contact wrenches $\bm{\lambda}$ to joint-space torques, and it is obtained by stacking the Jacobians which relate generalized velocities to limb end-effector motion as $\bm{J}_s = [\bm{J}_{C_1}^\top \quad \cdots \quad \bm{J}_{C_{n_c}}^\top]^\top$, with $n_c$ being the number of limbs in contact.
For fixed-base robots that are not subject to contact forces, we can simplify the equations of motion to
$\bm{M}(\bm{q})\bm{\dot{v}} + \bm{h}(\bm{q}, \bm{v}) = \bm{\tau}$.

In order to enforce the equations of motion of nonlinear systems, we define a set of equality constraints within our framework, the so-called \emph{defect constraints}.
Usually, these constraints are defined using a forward dynamics algorithm, but in this paper we argue that using inverse dynamics can be more computationally advantageous.

The standard problem of forward dynamics computes the joint accelerations resultant from commanding torques and applying forces to the robot at a given state, i.e.,
\begin{equation}
    \bm{\dot{v}}_{k}^{\ast} = f^\mathrm{fd}(\bm{q}_{k}, \bm{v}_{k}, \bm{\tau}_{k}, \bm{\lambda}_{k}),
    \label{equation:forward_dynamics}
\end{equation}
where $f^\mathrm{fd}(\cdot)$ is the function that solves forward dynamics.
The asterisk ${(\cdot)}^{\ast}$ denotes intermediately computed values,
whereas terms without an asterisk are \gls{NLP} variables.
Using the semi-implicit Euler method as the integration scheme
and $h = (t_F - t_I) / N$ as the integration time step, we can compute the state of the robot after $h$ seconds.
First, we integrate $\bm{\dot{v}}_{k}^{\ast}$ to compute the next generalized velocities
$\bm{v}_{k+1}^{\ast} = \bm{v}_{k} + h \, \bm{\dot{v}}_{k}^{\ast}.$
Then, we can compute the time derivative of the generalized coordinates, $\bm{\dot{q}}_{k+1}^{\ast}$, from those velocities, $\bm{v}_{k+1}^{\ast}$.
In turn, we integrate that time derivative to compute the next generalized coordinates,
$\bm{q}_{k+1}^{\ast} = \bm{q}_{k} + h \, \bm{\dot{q}}_{k+1}^{\ast}$.
After these calculations, we end up with two different values for the state of the system at mesh point $k+1$: one from the discretized \gls{NLP} variables, and another computed as a result of the controls applied to the system at mesh point $k$.
To enforce dynamical consistency, we define the defect constraints as
\begin{equation}
    \bm{q}_{k+1}^{\ast} - \bm{q}_{k+1} = \bm{0}
    \quad \text{and} \quad
    \bm{v}_{k+1}^{\ast} - \bm{v}_{k+1} = \bm{0}.
    \label{equation:defects_1}
\end{equation}

However, there is an alternative way to enforce dynamical consistency: with inverse dynamics.
In contrast to \eqref{equation:forward_dynamics}, inverse dynamics computes the joint torques and forces required to meet desired joint accelerations at a given state, i.e.,
\begin{equation}
    \bm{\tau}_{k}^{\ast} = f^\mathrm{id}(\bm{q}_{k}, \bm{v}_{k}, \bm{\dot{v}}_{k}^{\ast}, \bm{\lambda}_{k}),
    \label{equation:inverse_dynamics}
\end{equation}
where $f^\mathrm{id}(\cdot)$ is the function that solves the inverse dynamics problem,
and the desired joint accelerations can be calculated implicitly with
$\bm{\dot{v}}_{k}^{\ast} = (\bm{v}_{k+1} - \bm{v}_{k}) / h$.
Similarly to the forward dynamics case, we compute $\bm{\dot{q}}_{k+1}^{\ast}$ from $\bm{v}_{k+1}$, and integrate it to compute the next generalized coordinates $\bm{q}_{k+1}^{\ast}$.
And finally, we define the dynamics defect constraints as
\begin{equation}
    \bm{q}_{k+1}^{\ast} - \bm{q}_{k+1}   = \bm{0}
    \quad \text{and} \quad
    \bm{\tau}_{k}^{\ast} - \bm{\tau}_{k} = \bm{0}.
    \label{equation:defects_2}
\end{equation}

Notice that the main difference between equations \eqref{equation:defects_1} and \eqref{equation:defects_2} is that forward dynamics enforces consistency of the generalized velocities whereas inverse dynamics enforces consistency of joint torques commanded to the system.

The main subject of this paper revolves around the two formulations explained above to enforce the nonlinear system dynamics: \emph{forward dynamics} vs. \emph{inverse dynamics}.
We developed our framework with both options in mind, and we are able to easily toggle between one approach and the other,
which was particularly useful for our experiments.

\section{Experiments and Results}
This section is organized into four subsections:
\begin{itemize}
    \item [A.] Compares the computation time and number of solver iterations required to find locally-optimal solutions;
    \item [B.] Evaluates the robustness of each approach as problem discretization gets more coarse (larger time steps);
    \item [C.] Analyzes the performance of each formulation for the minimization of a cost function; and finally,
    \item [D.] Shows hardware validation of the planned motions.
\end{itemize}

All evaluations were carried out in a single-threaded process on an Intel i7-6700K CPU with clock frequency fixed at \SI{4.0}{\giga\hertz}, and \SI{32}{\giga\byte} \SI{2133}{\mega\hertz} memory.
The framework we propose has been implemented in Julia~\cite{bezanson2017julia},
using the rigid-body dynamics library \gls{RBD.jl}~\cite{rigidbodydynamicsjl},
and the optimization library Knitro~\cite{byrd2006knitro}.
To solve the formulated \gls{NLP} problems, we used the interior-point method of Waltz \textit{et al.} \cite{waltz2006interior}.

\subsection{Evaluation of Convergence}
In order to evaluate and compare forward dynamics against inverse dynamics in the context of direct transcription, we used our framework to specify tasks in the form of numerical optimization problems for different types of robots: a manipulator, a quadruped, and a humanoid.
Those robots were selected as they allows us to evaluate the formulations for distinct features: fixed- and floating-base systems, single-point and surface contacts, and low and high dimensionality.
For each task on each robot, we solved the optimization problem twice:
first defining the defect constraints with forward dynamics, and then with inverse dynamics.
The \emph{only} changing factor was the toggling between forward and inverse dynamics for the definition of the defect constraints;
every other aspect of the formulation was kept unchanged.

The performance of general \gls{NLP} solvers is greatly affected by the linear solver used for solving the linear systems of equations of the problem.
For this reason, we tested different state-of-the-art linear solvers exhaustively.
For interior-point methods, another important factor that affects performance is the update strategy of the barrier parameter.
Therefore, for all of our evaluations, we tested the different strategies available within the Knitro~\cite{byrd2006knitro} library exhaustively.

\begin{figure}[t]
    \captionsetup{font=small}
    \begin{subfigure}[t]{0.333\linewidth}
        \includegraphics[width=0.98\linewidth,center]{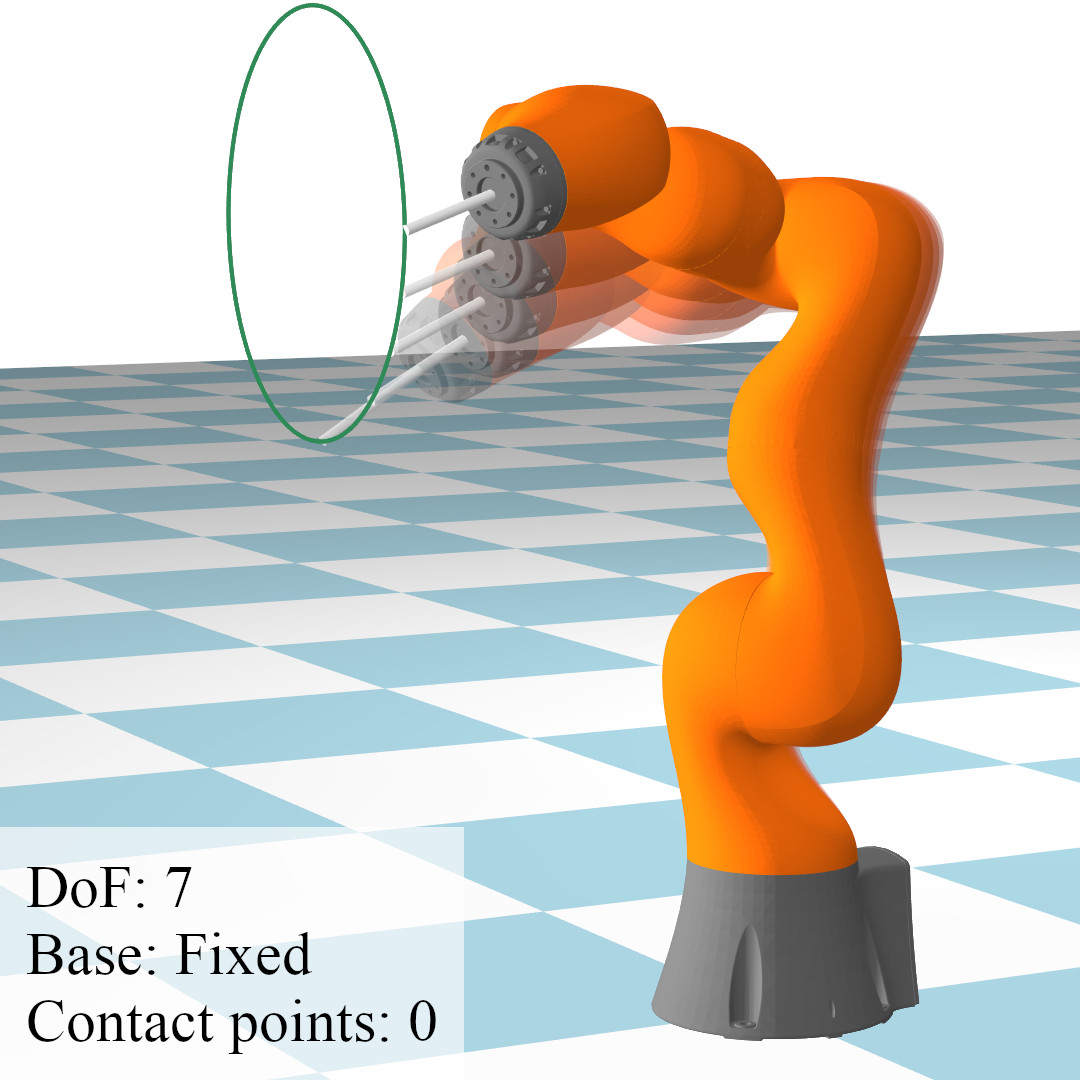}
        \caption{Manipulator}\label{figure:robot_kuka}
    \end{subfigure}\hfill%
    \begin{subfigure}[t]{0.333\linewidth}
        \includegraphics[width=0.98\linewidth,center]{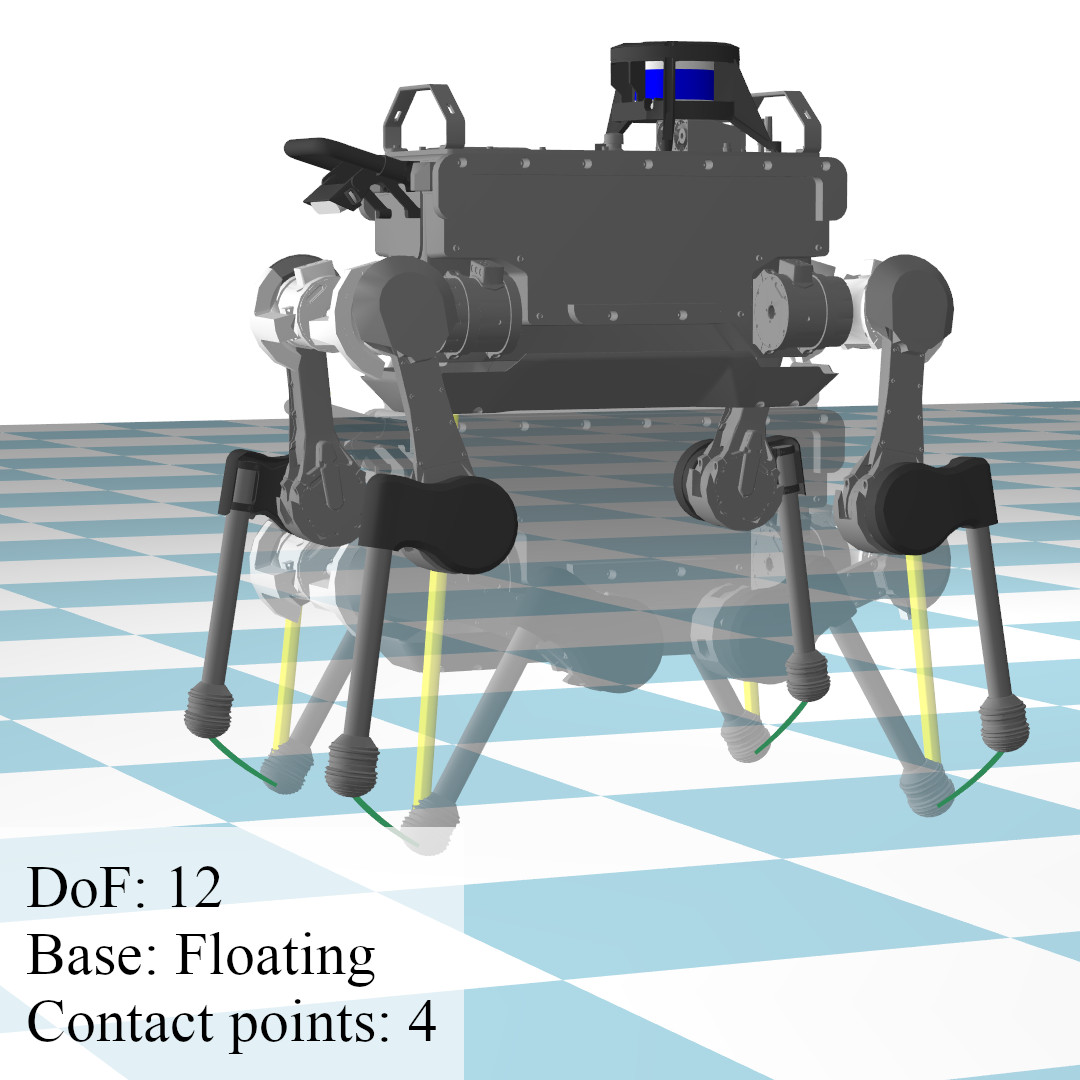}
        \caption{Quadruped}\label{figure:robot_anymal}
    \end{subfigure}\hfill%
    \begin{subfigure}[t]{0.333\linewidth}
        \includegraphics[width=0.98\linewidth,center]{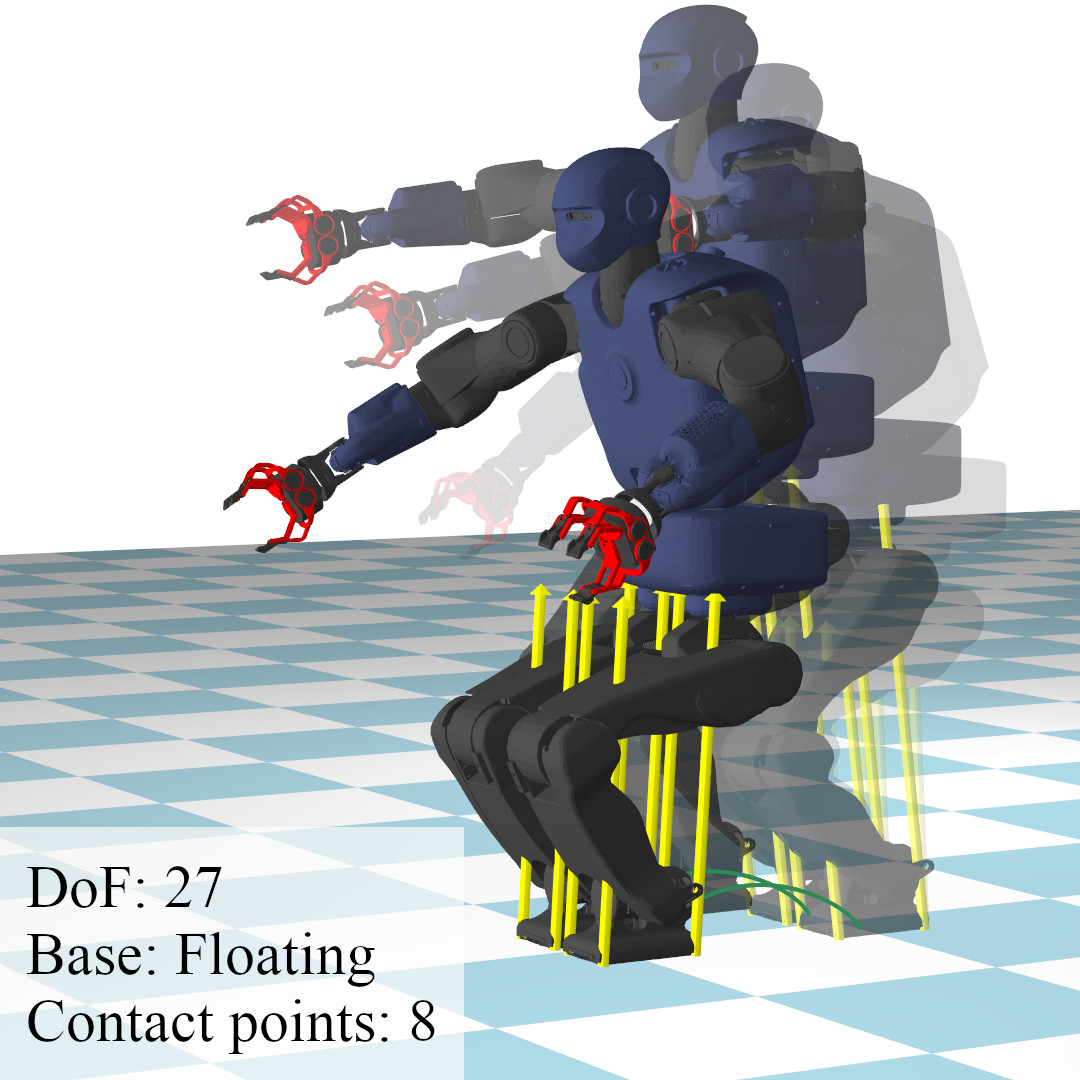}
        \caption{Humanoid}\label{figure:robot_talos}
    \end{subfigure}
    \caption{
        Left to right: KUKA iiwa tracing a circular path, ANYmal jumping in-place, and TALOS jumping forward.
    }\label{figure:robot_tasks}
    \vspace{-12pt}
\end{figure}

In the remainder of this subsection, we present the task specifications for each robot and indicate all the parameters for reproducibility.
Then, we present the results we obtained for each task, which evaluated the solver's performance in terms of computation time and number of iterations taken by the solver until a locally-optimal solution was found.

\subsubsection{Manipulator}
We evaluated the different formulations using a fixed-base robot arm with seven \gls{DoF} (shown in \autoref{figure:robot_kuka}).
We specified the end-effector to trace a circular path given by
$[0.5, 0.2 \cos \theta, 0.8 + 0.2 \sin \theta] \forall \theta \in [0,2\pi]$.
The total duration was set to \SI{2.0}{\second} and the trajectory was discretized at \SI{150}{\hertz}, resulting in a total of \num{301} mesh points.

\subsubsection{Quadruped}
The quadruped robot we used is shown in \autoref{figure:robot_anymal}.
This system is more complex than the manipulator due to its floating-base, more \gls{DoF} (three motors per leg), and because it needs to handle contact forces.
We defined a jumping task by enforcing the contact forces to be zero for a short period of time.
The trajectory was discretized at \SI{100}{\hertz}, the total duration of the motion was \SI{2.0}{\second}, and the interval specified for the flight-phase was $[1.0,1.2]$ \si{\second}.
We did not constrain feet positions during the flight-phase, which allowed the solver to converge to a solution where the feet swing most naturally according to the system dynamics.

\subsubsection{Humanoid}
Finally, we considered the humanoid robot shown in \autoref{figure:robot_talos}.
This robot is more complex than the quadruped because it has 27 \gls{DoF}\footnote{%
    The real robot has more \gls{DoF}: grippers, neck, and one more \gls{DoF} at the torso. For simplicity, we assumed those joints were fixed to zero.
} (seven per arm, six per leg, and one at the torso), and its feet cannot be simplified to single-point contacts.
We also defined a jumping task for this robot.
The motion duration was \SI{1.2}{\second}, discretized at \SI{125}{\hertz}, and the interval for the flight-phase was $[0.5,0.8]$ \si{\second}.

For all the tasks, the initial guess was a fixed standing configuration and zero velocities, torques, and contact forces.

The results of the experiments on these robots are shown in \autoref{table:results},
where smaller numbers indicate better performance.
The rows of the table are grouped according to robot, dynamics, and linear solver.
Each row shows the time taken to solve the optimization problem for each barrier update strategy,
as well as the total number of iterations (within parenthesis).
The last column shows the time spent on each iteration, averaged over all of the update strategies.

\begin{table}[t]
    \captionsetup{font=small}
    \centering
    \caption[Computation time and number of iterations for each robot's task.]{
        Computation time\footnotemark (in seconds) and number of iterations (within parenthesis) for each robot.
        The best computation time for each dynamics and each robot is highlighted in bold.
    }\label{table:results}
    \begin{adjustbox}{width=1\linewidth}
        \begin{tabular}{@{\hskip 0pt}c@{\hskip 8pt}ccr@{\hskip 0pt}rr@{\hskip 0pt}rr@{\hskip 0pt}rc@{\hskip 0pt}}
            \toprule
                                                                   &                                                  & Linear & \multicolumn{6}{c}{Barrier strategy (\texttt{bar\_murule})}                                                        & Average time                 \\
                                                                   &                                                  & Solver & \multicolumn{2}{c}{\texttt{adaptiv}} & \multicolumn{2}{c}{\texttt{dampmpc}} & \multicolumn{2}{c}{\texttt{quality}} & per iteration (\si{\second}) \\ \midrule[\heavyrulewidth]
            \multirow{6.5}{*}{\begin{turn}{90}KUKA iiwa\end{turn}} & \multirow{3}{*}{\begin{turn}{90}Fwd D\end{turn}} & MA27   &                 \textbf{0.40} &  (5) &                         0.49  &  (6) &                         0.43  &  (5) &             0.08 $\pm$ 0.002 \\
                                                                   &                                                  & MA57   &                         0.41  &  (5) &                         0.48  &  (6) &                         0.42  &  (5) &             0.08 $\pm$ 0.001 \\
                                                                   &                                                  & MA97   &                         0.46  &  (5) &                         0.56  &  (6) &                         0.50  &  (5) &             0.09 $\pm$ 0.002 \\ \cdashlinelr{2-10}
                                                                   & \multirow{3}{*}{\begin{turn}{90}Inv D\end{turn}} & MA27   &                 \textbf{0.20} &  (4) &                         0.26  &  (5) &                         0.23  &  (4) &             0.05 $\pm$ 0.002 \\
                                                                   &                                                  & MA57   &                         0.22  &  (4) &                         0.28  &  (5) &                         0.24  &  (4) &             0.05 $\pm$ 0.001 \\
                                                                   &                                                  & MA97   &                         0.27  &  (4) &                         0.33  &  (5) &                         0.29  &  (4) &             0.07 $\pm$ 0.002 \\ \midrule[\heavyrulewidth]
            \multirow{6.5}{*}{\begin{turn}{90}ANYmal B\end{turn}}  & \multirow{3}{*}{\begin{turn}{90}Fwd D\end{turn}} & MA27   &                         2.26  &  (7) &                         2.80  &  (9) &                         2.34  &  (7) &             0.31 $\pm$ 0.021 \\
                                                                   &                                                  & MA57   &                         2.80  & (10) &                         2.49  &  (9) &                        32.13  & (99) &             0.29 $\pm$ 0.020 \\
                                                                   &                                                  & MA97   &                 \textbf{2.21} &  (7) &                         2.76  &  (9) &                         2.26  &  (7) &             0.31 $\pm$ 0.009 \\ \cdashlinelr{2-10}
                                                                   & \multirow{3}{*}{\begin{turn}{90}Inv D\end{turn}} & MA27   &                         2.99  & (13) &                         2.60  & (11) &                         2.18  &  (9) &             0.23 $\pm$ 0.005 \\
                                                                   &                                                  & MA57   &                         2.90  & (13) &                         2.54  & (11) &                 \textbf{2.10} &  (9) &             0.22 $\pm$ 0.004 \\
                                                                   &                                                  & MA97   &                         3.21  & (13) &                         2.81  & (11) &                         2.37  &  (9) &             0.25 $\pm$ 0.007 \\ \midrule[\heavyrulewidth]
            \multirow{6.5}{*}{\begin{turn}{90}TALOS\end{turn}}     & \multirow{3}{*}{\begin{turn}{90}Fwd D\end{turn}} & MA27   &                          ---  &      &                          ---  &      &                        14.82  & (15) &             0.94 $\pm$ 0.115 \\
                                                                   &                                                  & MA57   &                          ---  &      &                        13.47  & (19) &                        44.23  & (66) &             0.68 $\pm$ 0.020 \\
                                                                   &                                                  & MA97   &                        13.42  & (18) &                \textbf{11.48} & (15) &                        12.92  & (16) &             0.76 $\pm$ 0.026 \\ \cdashlinelr{2-10}
                                                                   & \multirow{3}{*}{\begin{turn}{90}Inv D\end{turn}} & MA27   &                         8.38  & (15) &                         8.02  & (13) &                        38.59  & (70) &             0.57 $\pm$ 0.035 \\
                                                                   &                                                  & MA57   &                         7.36  & (15) &                \textbf{ 6.59} & (13) &                        36.88  & (73) &             0.50 $\pm$ 0.014 \\
                                                                   &                                                  & MA97   &                         7.43  & (15) &                         6.61  & (13) &                        41.84  & (82) &             0.50 $\pm$ 0.012 \\
            \bottomrule
        \end{tabular}
    \end{adjustbox}
    \vspace{-12pt}
\end{table}

\footnotetext{%
    This table (and future tables) show the minimum time value observed over \num{10} trials.
    Reporting the minimum time is more reliable than the median or the mean, since all measured noise is positive, as explained in \cite{chen2016robust}.
}

In general, we can see that the computation time depends mostly on the complexity of the system, regardless of linear solver or barrier update strategy;
i.e., solving the manipulator task was faster than solving the quadruped task, which in turn was faster than the humanoid task.
More importantly, for each robot and given the same choice of linear solver and update strategy, the computation time of inverse dynamics was better than forward dynamics.
We can also see that the number of iterations required to solve the problem did not change significantly (apart from a few exceptions).
This indicates that the difficulty of the problem itself did not change with the different dynamics defects; it just took longer to solve using forward dynamics---as supported by the information in the last column of the table.

\subsection{Robustness to Coarser Problem Discretizations}
\begin{figure}[t]
    \captionsetup{font=small}
    \centering
    \includegraphics[width=\linewidth]{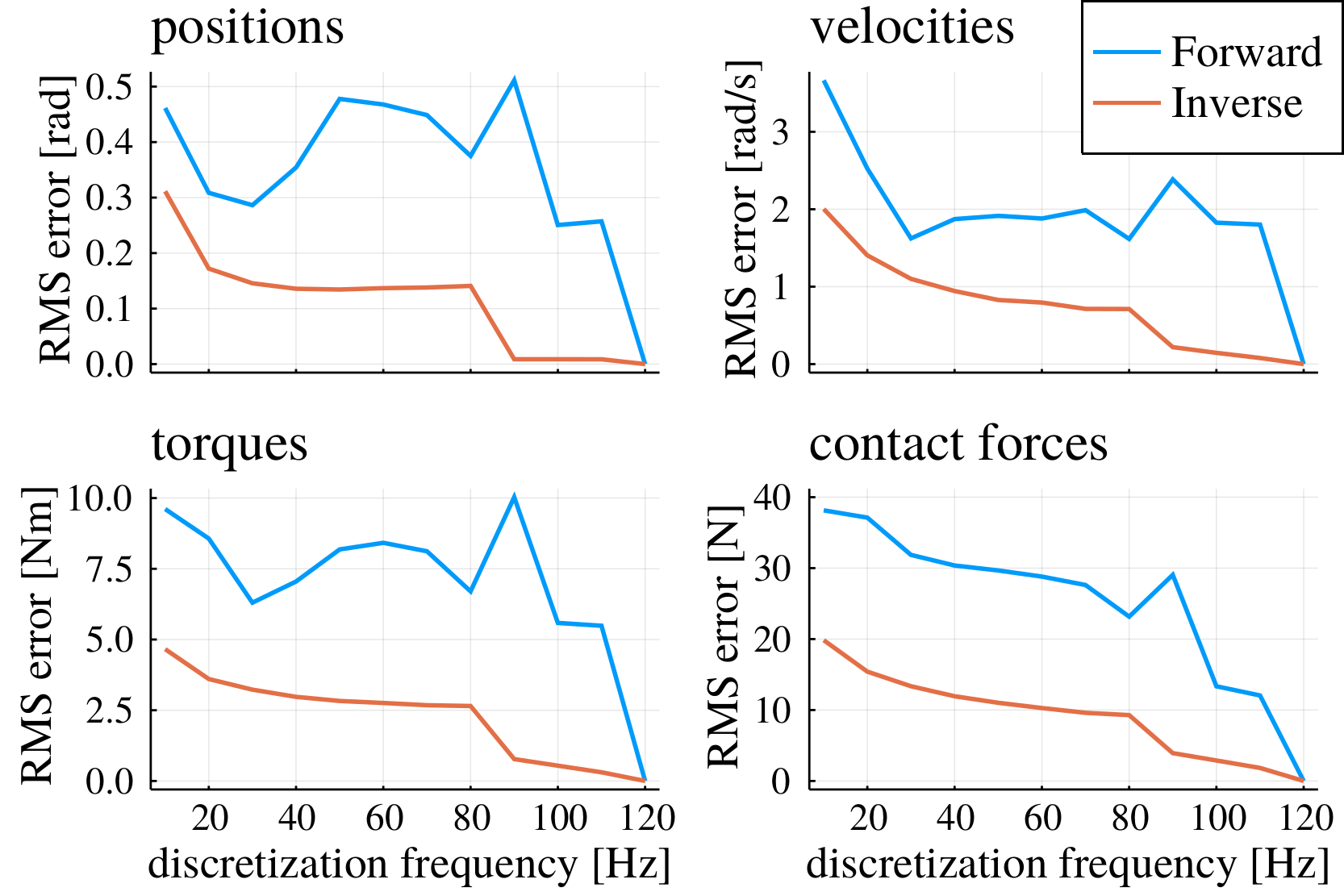}
    \caption{
        Root-mean-square error (RMSE) of joint positions, velocities, torques, and contact forces of each formulation for different discretizations, using a baseline of \SI{120}{\hertz}.
    }\label{figure:rms_error}
    \vspace{-16pt}
\end{figure}

In the next experiment, we compare the ability of each formulation to handle trajectories discretized using fewer mesh points.
We defined the same quadruped jumping task repeatedly, but transcribed it with different resolutions.
First, we divided the trajectory into equally spaced segments with a time step of $h=0.01$; we solved the optimization problem and took the resulting trajectory as our baseline.
Then, we incrementally changed $h$, making the problem more coarse each time, and compared the obtained trajectories against the baseline.
The problems were initialized with a nominal configuration repeated for each point, and zero velocities, torques and contact forces.
The results of this experiment are shown in \autoref{figure:rms_error} and \autoref{table:stress_test}.

The plots in \autoref{figure:rms_error} show that the solutions deviate more from the baseline as the number of mesh points used to discretize the problem decreases (in the $x$-axis, from right to left).
But more importantly, the plots reveal that the rate at which deviation occurs is significantly different depending on the formulation.
We can see that the \gls{RMSE} of the formulation using inverse dynamics is significantly lower than that of the forward.

\begin{table}[t]
    \captionsetup{font=small}
    \centering
    \caption{Computation time and number of iterations required by different problem discretizations, for the quadruped jump.}
    \label{table:stress_test}
    \begin{tabular}{ccccccc}
        \toprule
        \multirow{2}{*}{Frequency} &  & \multicolumn{2}{c}{Forward Dyn.} &       & \multicolumn{2}{c}{Inverse Dyn.}                               \\
                                   &  & Time (\si{\second})              & Iter. &                                  & Time (\si{\second}) & Iter. \\ \midrule
        \SI{100}{\hertz}           &  & 3.289                            & 10    &                                  & 3.295               & 13    \\
        \SI{90}{\hertz}            &  & 3.916                            & 14    &                                  & 2.774               & 13    \\
        \SI{80}{\hertz}            &  & 5.227                            & 21    &                                  & 1.821               & 9     \\
        \SI{70}{\hertz}            &  & 4.062                            & 19    &                                  & 1.344               & 8     \\
        \SI{60}{\hertz}            &  & 1.518                            & 9     &                                  & 1.193               & 8     \\
        \SI{50}{\hertz}            &  & 2.226                            & 16    &                                  & 0.983               & 8     \\
        \SI{40}{\hertz}            &  & 1.099                            & 9     &                                  & 0.785               & 8     \\
        \SI{30}{\hertz}            &  & 0.731                            & 8     &                                  & 0.534               & 7     \\
        \SI{20}{\hertz}            &  & 0.433                            & 7     &                                  & 0.354               & 7     \\
        \bottomrule
    \end{tabular}
\end{table}

\autoref{table:stress_test} shows the computation time (in seconds) and the number of iterations required to solve the quadruped task using different discretizations.
We can see that the time required to solve the problem using inverse dynamics follows a clear pattern: it decreases as the problem gets more coarse; and the same goes for the number of iterations.
In contrast, a pattern does not seem to exist for forward dynamics.

The results shown in \autoref{figure:rms_error} and \autoref{table:stress_test} provide strong evidence that
defining the defect constraints with inverse dynamics is the approach more robust to different problem discretizations,
both in terms of deviation from realistic solutions and in terms of computation performance.

\begin{figure*}[t]
    \captionsetup{font=small}
    \begin{subfigure}[t]{0.2\linewidth}
        \includegraphics[width=0.98\linewidth,center]{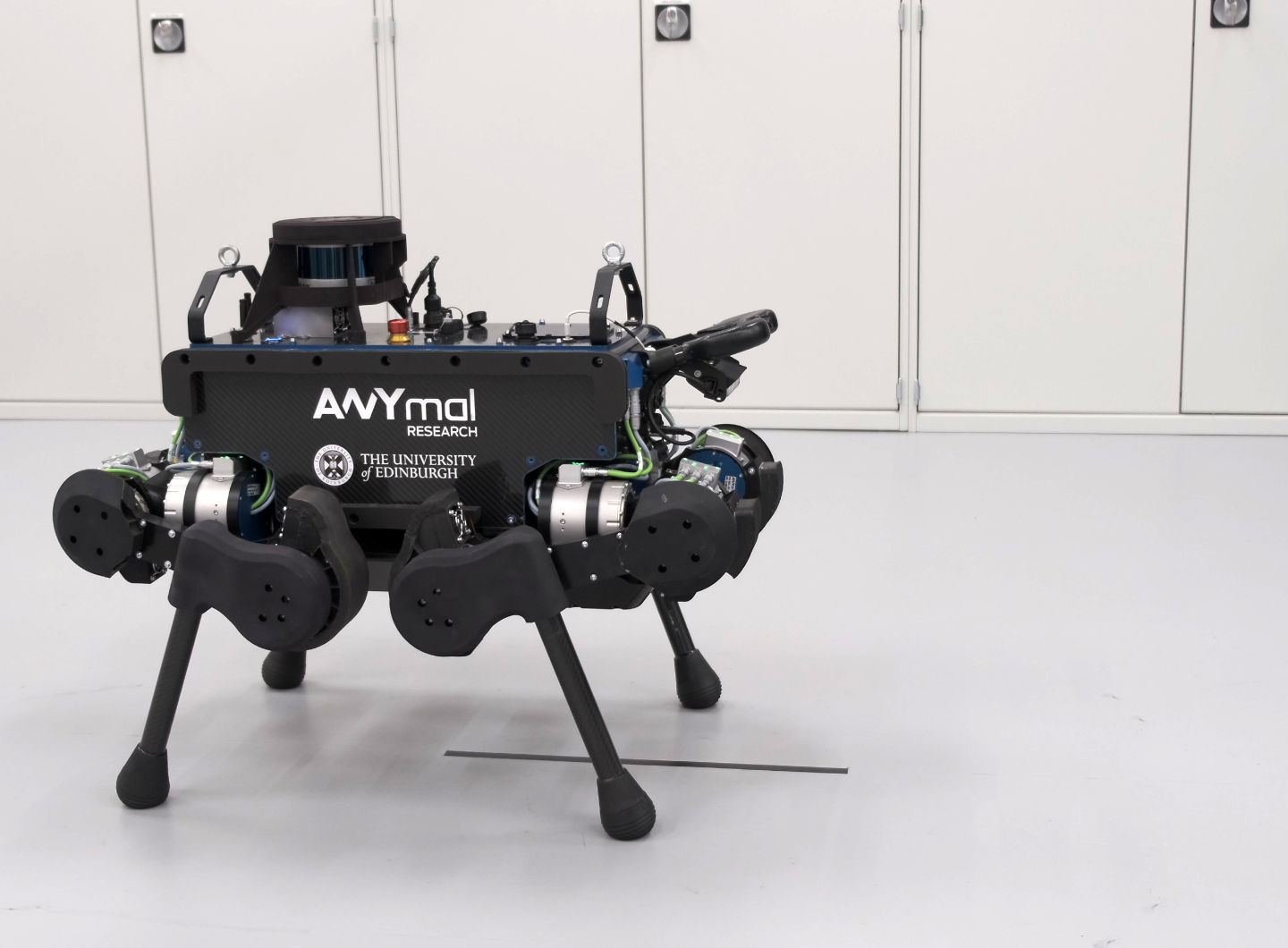}
        \caption{Initial configuration}
    \end{subfigure}\hfill%
    \begin{subfigure}[t]{0.2\linewidth}
        \includegraphics[width=0.98\linewidth,center]{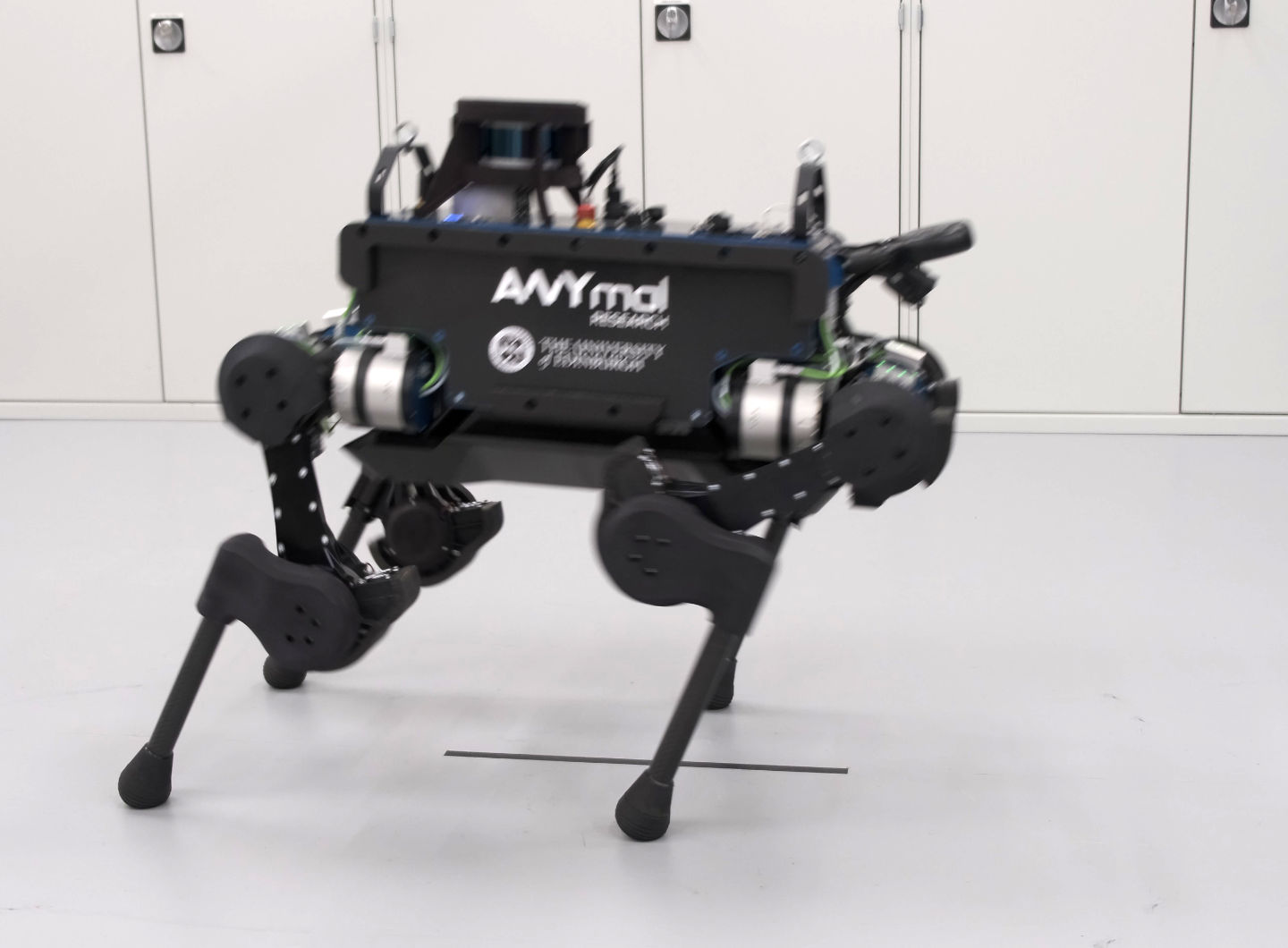}
        \caption{Take-off}
    \end{subfigure}\hfill%
    \begin{subfigure}[t]{0.2\linewidth}
        \includegraphics[width=0.98\linewidth,center]{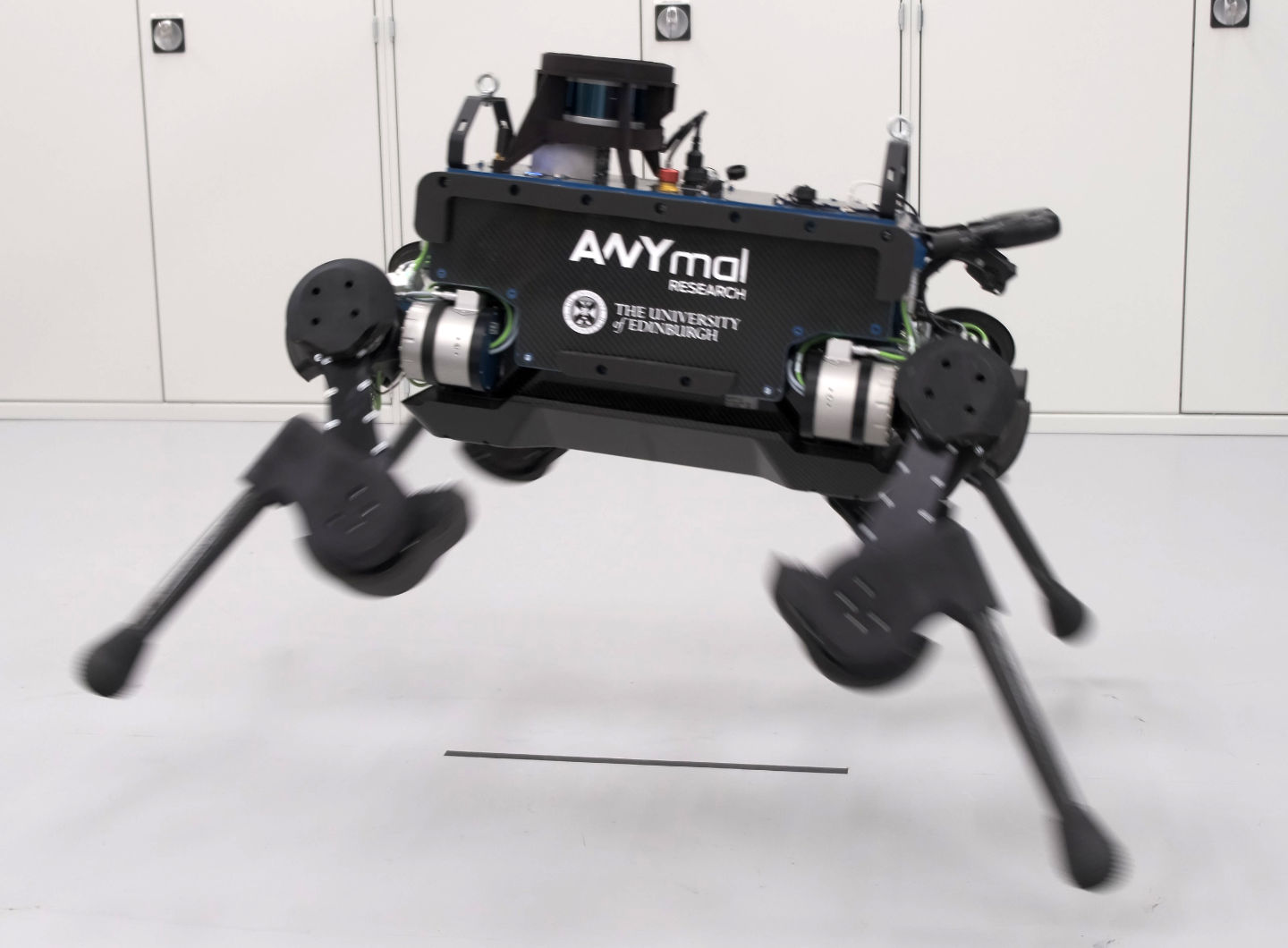}
        \caption{Full-flight phase}
    \end{subfigure}\hfill%
    \begin{subfigure}[t]{0.2\linewidth}
        \includegraphics[width=0.98\linewidth,center]{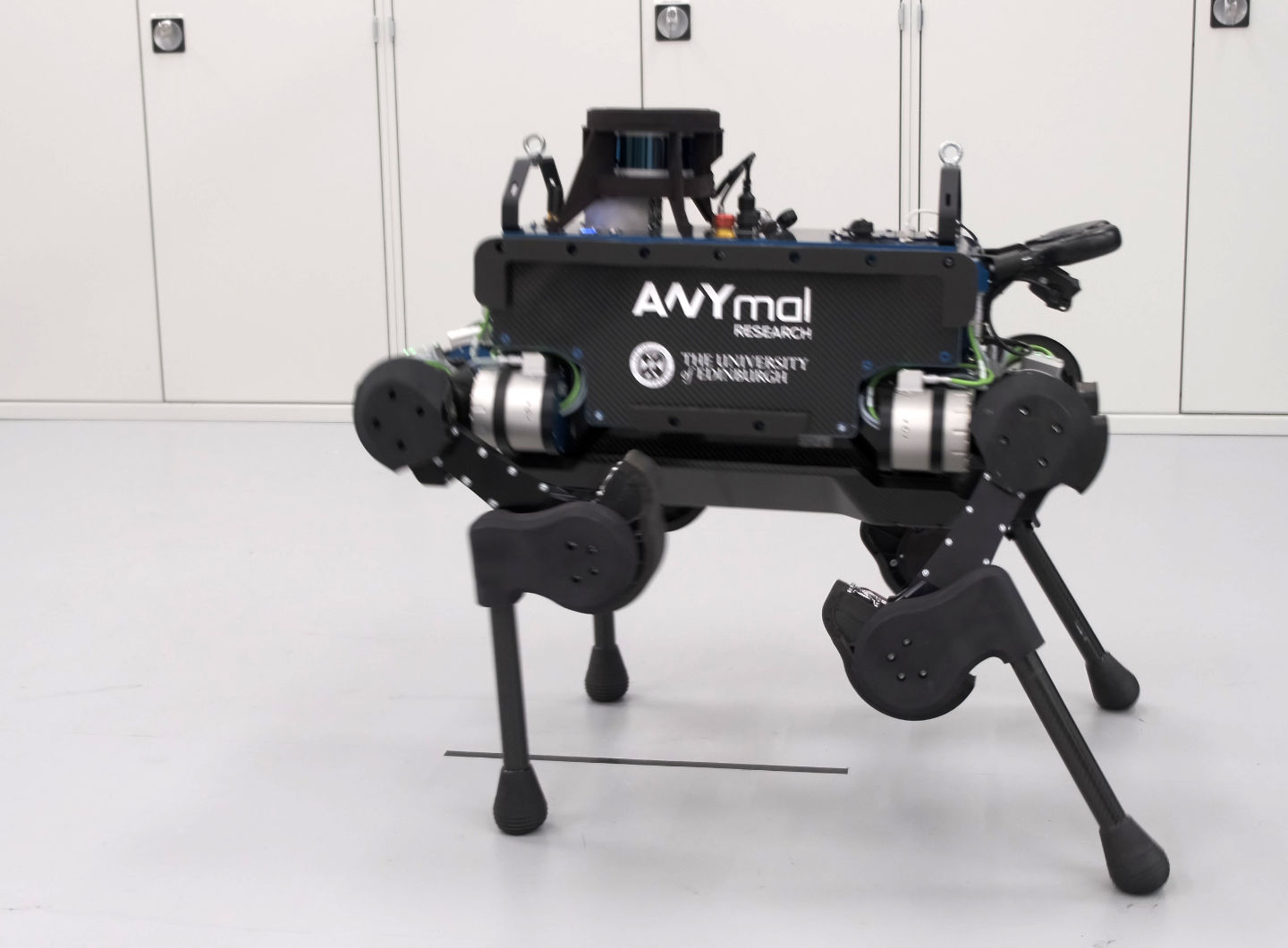}
        \caption{Landing}
    \end{subfigure}\hfill%
    \begin{subfigure}[t]{0.2\linewidth}
        \includegraphics[width=0.98\linewidth,center]{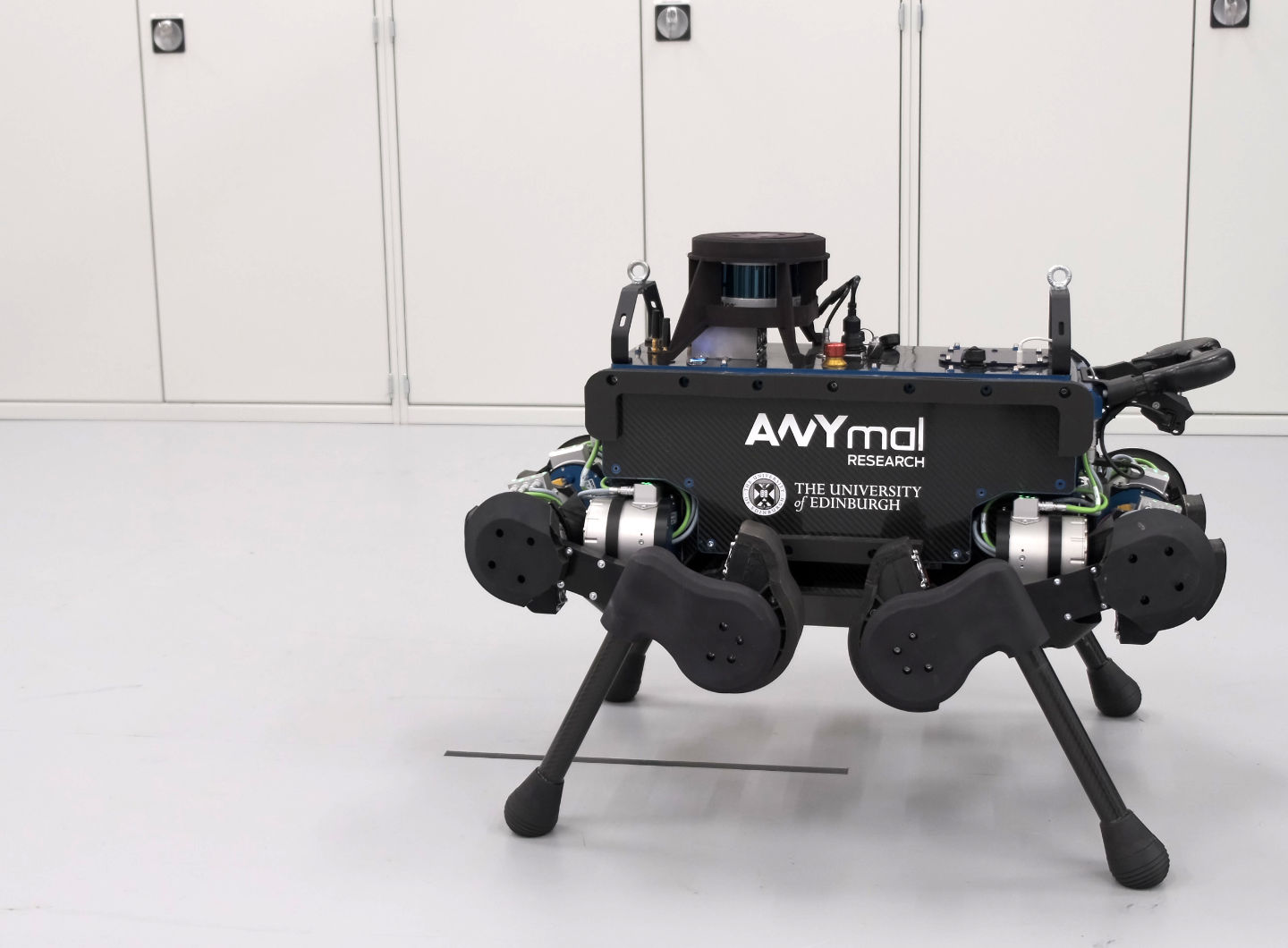}
        \caption{Final configuration}
    \end{subfigure}
    \caption{
        Snapshots of ANYmal~\cite{hutter2017anymal} performing a \SI{0.5}{\meter}-long jump.
        The length of the black tape on the ground is \SI{0.5}{\meter}.
    }\label{figure:snapshots_anymal}
    \vspace{-8pt}
\end{figure*}

\subsection{Optimization with an Objective Function}
Thus far we have analyzed the trajectory optimization performance for feasibility problems.
However, in optimization, it is common to define a cost function to be minimized (or a value function to be maximized).
In this next experiment, we evaluate the performance of our formulation when a cost function is considered.
We minimize the actuator torques and ground-reaction contact forces with $\min_{\bm{\xi}} \quad \sum_{k=1}^{M-1} \frac{\bm{\tau}_k^\top \bm{\tau}_k + \bm{\lambda}_k^\top \bm{\lambda}_k}{M-1}$.
We tested this cost function on the quadruped jump task, with the MA57 linear solver and \texttt{adaptive} barrier parameter update strategy.

Both formulations converged to very similar solutions: the \gls{RMSE} between the two trajectories was \num{0.038}.
The final objective value was \num{3.567801e+04} and \num{3.567804e+04} for forward and inverse dynamics, respectively.
Despite converging to similar solutions, the formulation employing inverse dynamics finished in \SI{6.208}{\second}, showing better performance than the formulation using forward dynamics which took \SI{14.570}{\second}.
The time in seconds corresponds to the minimum value measured over a total of 10 samples.

\begin{figure}[t]
    \captionsetup{font=small}
    \centering
    \includegraphics[width=\linewidth]{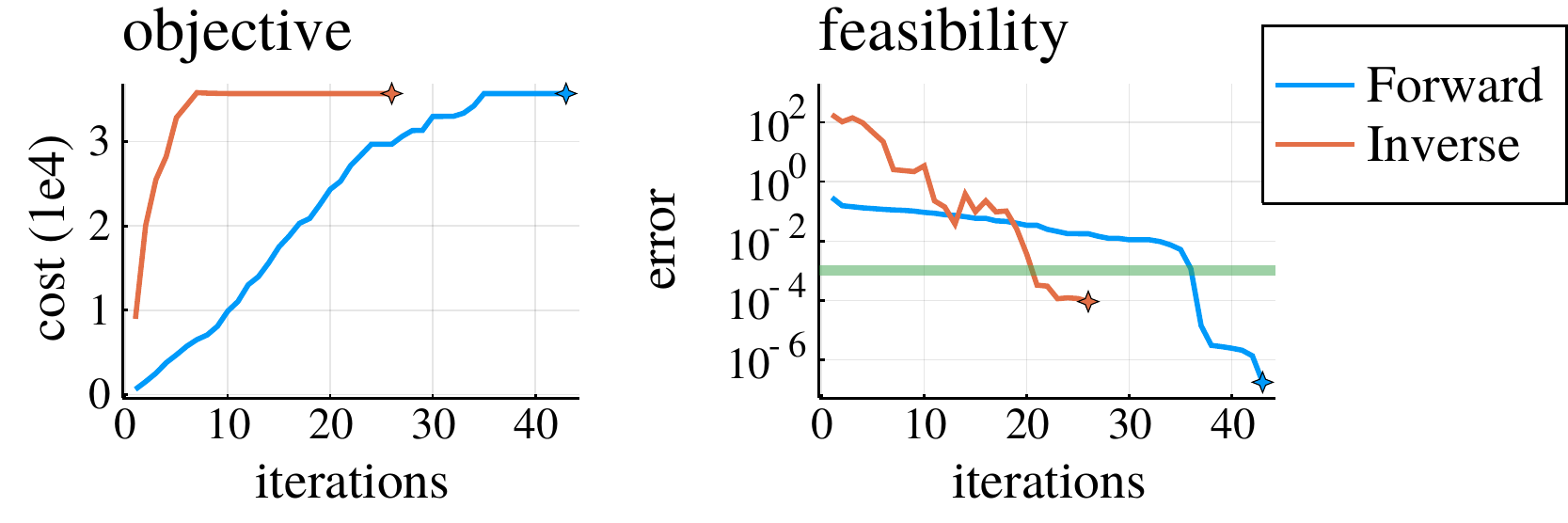}
    \caption{
    Evolution of the cost and the feasibility error during convergence.
    The faint-green line at $y=10^{-3}$ denotes the tolerance under which we consider a problem to be feasible.
    }\label{figure:minimization}
    \vspace{-12pt}
\end{figure}

\autoref{figure:minimization} shows the evolution of the cost and feasibility error throughout the optimization.
The star-shaped marker denotes the point at which the local minimum of the problem was found.
In the left plot, we can see that inverse dynamics reached values close to the optimal cost much earlier than forward dynamics.
In the right plot, we can see that inverse dynamics required less iterations than forward dynamics to cross the faint-green line, which marks the point at which the error becomes acceptable to be considered feasible.
Inverse dynamics converged in \num{26} iterations, and forward dynamics converged in \num{43} iterations.
Inadvertently, one advantage of the forward formulation was that its final feasibility error was smaller than that of inverse dynamics.

\subsection{Hardware Validation}
We conducted real-world experiments with ANYmal~\cite{hutter2017anymal} and TALOS~\cite{stasse2017talos} to validate the trajectories computed with our framework.
The motion planning is performed offline and then the trajectories are sent to the controller for playback.
To execute the whole-body motions, we commanded each joint with feedforward torque and feedback on joint position and velocity.
For the quadruped, we updated the references for each joint's position, velocity, and torque at \SI{400}{\hertz}.
The decentralized motor controller at every joint closes the loop compensating for friction effects.
On the humanoid, we updated the references at \SI{2}{\kilo\hertz}, and a centralized controller compensates for the motor dynamics and friction.

\autoref{figure:cover} and \autoref{figure:snapshots_anymal} contain snapshots of the jumps realized with the humanoid and with the quadruped, respectively.
These experiments can be seen in our supplementary video:
\texttt{\url{https://youtu.be/pV4s7hzUgjc}}.
Jumping motion is challenging to execute in real hardware because it includes a severely underactuated phase when the robot is fully off the ground.
Nonetheless, our controller is able to execute our planned trajectories reliably, attesting the dynamical consistency of our formulation.

\section{Discussion}
\label{sec:discussion}

The results of this work indicate that direct transcription implementations relying on forward dynamics to define the defect constraints can be reformulated with inverse dynamics to see an increase in performance, for both feasibility or minimization problems, and without sacrificing the feasibility of the solutions to the optimization problem.
An additional reason to prefer inverse dynamics is robustness to coarser discretizations, both in terms of computation efficiency and faithfulness of solutions with respect to finer discretizations.

When minimizing a cost function, the locally-optimal solutions computed with either formulation are essentially the same.
However, when an objective function is not considered, the formulations may diverge to different solutions.
Experimentally, we have observed that the solutions computed with inverse dynamics are easier to perform in real hardware.
The reasons behind this divergence are not yet clear to us, and this is something we plan to investigate in future work.

Erez and Todorov \cite{erez2012trajectory} observed a striking feature in their results: an emergent coordination between legs and opposite arms during a running gait.
In this work, for the humanoid jumping task, we also observed such emerging behavior:
the resulting motions swing the arms upwards to build-up energy before the take-off instant.
Both in \cite{erez2012trajectory} and our work, these features originated without any explicit modeling---reaffirming the power of dynamic trajectory optimization.

In recent work \cite{ferrolho2020optimizing}, we took into account uncertainty and robustness to disturbances using direct transcription.
Considering uncertainty usually incurs additional computational cost due to more complex problem formulations.
With the findings from this paper, we plan to redefine the dynamics defect constraints in that work with inverse dynamics, improving the performance of our robustness framework and making it more competitive.

\bibliographystyle{IEEEtran}
\bibliography{IEEEabrv,references}

\end{document}